\pdfoutput=1

\documentclass[letterpaper, 10 pt, conference]{ieeeconf}  
\usepackage[ruled,vlined]{algorithm2e}
\usepackage{amssymb}
\usepackage{dsfont}
\usepackage{mathtools}
\usepackage[english]{babel}
\usepackage{hyperref}

\usepackage[final]{graphicx}
\usepackage{subcaption}

\newtheorem{lemma}{Lemma}

\IEEEoverridecommandlockouts                              

\title{\LARGE \bf
Feeling Optimistic? Ambiguity Attitudes for Online Decision Making
}

\author{Jared J. Beard$^{1}$, R. Michael Butts$^{2}$, and Yu Gu$^{3}$ 
\thanks{$^{1}$Jared J. Beard is a Doctoral Candidate with the Department of Mechanical and Aerospace engineering at WVU$^\dagger$
        {\tt\small jbeard6@mail.wvu.edu}}%
\thanks{$^{2}$ R. Michael Butts is a Master's student with the Lane Department of Computer Science and Electrical Engineering at WVU$^\dagger$
        {\tt\small rmb0034@mix.wvu.edu}}%
\thanks{$^{3}$Yu Gu is faculty with the Department of Mechanical and Aerospace engineering at WVU$^\dagger$
        {\tt\small yu.gu@mail.wvu.edu}}%
\thanks{$^{\dagger}$ West Virginia University, Morgantown, WV 26506, USA}%
}

\begin{document}

\maketitle
\thispagestyle{empty}
\pagestyle{empty}

\begin{abstract}
Due to the complexity of many decision making problems, tree search algorithms often have inadequate information to produce accurate transition models. This results in ambiguities (uncertainties for which there are multiple plausible models). Faced with ambiguities, robust methods have been used to produce safe solutions\textemdash often by maximizing the lower bound over the set of plausible transition models. However, they often overlook how much the representation of uncertainty can impact how a decision is made. This work introduces the Ambiguity Attitude Graph Search (AAGS), advocating for more comprehensive representations of ambiguities in decision making. Additionally, AAGS allows users to adjust their ambiguity attitude (or preference), promoting exploration and improving users' ability to control how an agent should respond when faced with a set of plausible alternatives. Simulation in a dynamic sailing environment shows how environments with high entropy transition models can lead robust methods to fail. Results further demonstrate how adjusting ambiguity attitudes better fulfills objectives while mitigating this failure mode of robust approaches. 
Because this approach is a generalization of the robust framework, these results further demonstrate how algorithms focused on ambiguity have applicability beyond safety-critical systems. 

\end{abstract}


\section{Introduction}

As autonomous agents navigate more complex environments, they are confronted with increasing model uncertainty. 
This is especially important with Monte Carlo tree search techniques, such as upper confidence trees (UCT) \cite{kocsis2006bandit}. Tree search algorithms sample a blackbox simulator to produce transition models for each state encountered. However, many of these algorithms will be inaccurate due to limitations on sampling and the tree structure of the graph \cite{leurent2020monte}. Consider a sailing robot, many diverse encounters and long operational periods are required before enough data is present to model commonly occurring conditions at sea. This leaves robots with too little information to reliably predict outcomes\textemdash that is to say the uncertainty is non-measurable \cite{bevan2022ambiguities} or ambiguous. Conditioned on some prior information, these ambiguities can be thought of as equally valid models of a system. For example, when sailing, we cannot precisely measure the conditions beyond the immediate area. Without this distinguishing information, we cannot predict precisely where the boat will end up prior to acting.
\begin{figure}[h]
    \centering
    \includegraphics[width=0.98\linewidth]{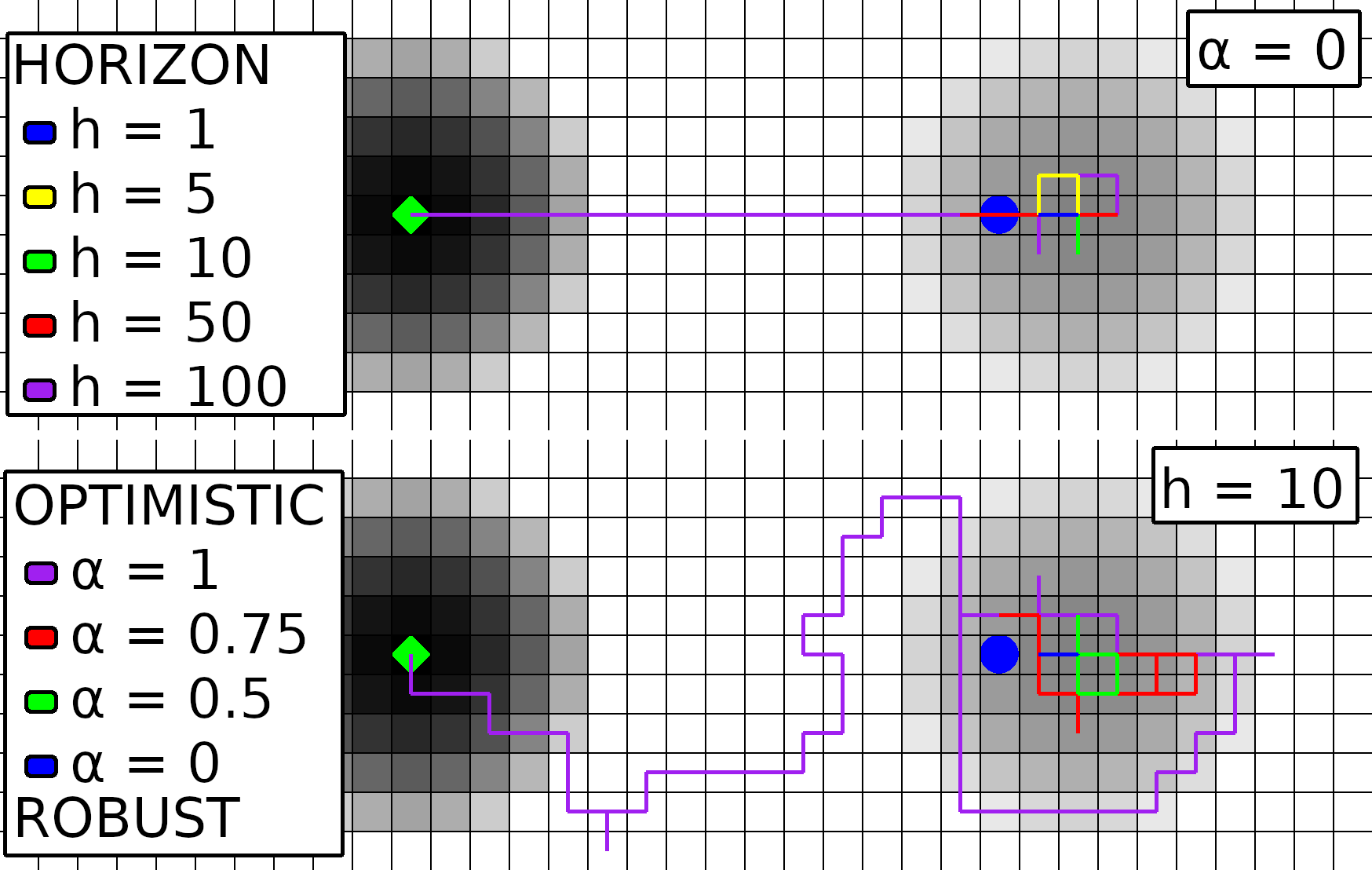}
    \caption{Suppose an agent (blue) needs to reach a goal (green). The agent is penalized for time, but receives increasing reward in darker cells. When conducting tree search, each node will have a limited number of samples to approximate the transition model. From these samples, multiple equally valid models may be inferred, resulting in ambiguity. Robust methods assume this ambiguity should be avoided and aim to improve the worst-case outcomes. This may require long planning horizons to escape local minima and reach the goal (top). Conversely, agents with more optimistic attitudes, as this work demonstrates, seek ambiguity to explore their environment and reach goals beyond their planning horizon (bottom).} 
    \label{fig:tunnel_top}
\end{figure}
Failure to consider these ambiguities in the context of decision making can distort the perception of uncertainty and leaves robots oblivious to potential fault conditions. 

Despite an extensive body of literature about ambiguities \cite{cuzzolin2020geometry,bevan2022ambiguities}, tools for decision making under ambiguity are notably lacking \cite{keith2021survey}. Instead tools typically assume away this type of uncertainty or address it implicitly. In the case of Bayesian decision making, for example, a maximum \textit{a posteriori} (MAP) estimate is used to represent transition models \cite{mern2021bayesian}. Such methods do not consider how probable or accurate the MAP models are, which can be problematic with limited data. Alternatively, variants of upper confidence trees use statistically derived rules or bounds such as UCB1 \cite{kocsis2006bandit} to approximate the value of an action. Again, the accuracy of models remains unclear, and these representations obscure the impact of tuning parameters in affecting the policy. Expressions of ambiguity, on the other hand, present clear descriptions of what is known about the model and how confidently mass can be assigned to transition outcomes.

Some more recent works consider ambiguity in a limited sense. These approaches generally fit under the auspices of robust decision making. Robust decision making tools construct a set of models and seek to maximize the lower bound on a state's value \cite{iyengar2005robust,wiesemann2013robust}. Representations of ambiguity in this context are limited to the use of confidence intervals \cite{leurent2020monte, jonsson2020planning} or chance constraints on undesirable outcomes \cite{rostov2021robust}. These representations neglect other information conveyed by the transition model. For example, by considering the distribution of outcomes more explicitly, we can arrive at more informed decisions \cite{bellemare2017distributional}; coupled with representations of ambiguity we can better assess the knowledge carried by a decision maker. Additionally, the conservatism of robust methods limits their use to safety critical systems. In many other applications (\textit{e.g.}, service robots), giving decision makers the option to trade-off between robust policies and optimistic policies when faced with ambiguity can lead to more desirable outcomes (Fig. \ref{fig:tunnel_top}). 

\subsection{Contributions}
This work contributes the following:
\begin{itemize}
    \item Ambiguity attitudes are introduced to online decision making through the novel ambiguity attitude graph search (AAGS) algorithm. This is demonstrated to generalize robust decision making and incorporates exploratory behaviors directly into an agent's decision making process.
    \item A technique for evaluating sets of transition models from confidence intervals is developed. This technique maps confidence intervals to ambiguous probability mass functions known as belief functions. 
    \item A novel approximation of the relationship between the number of samples, model accuracy, and model confidence is demonstrated. This is intended to be used in place of bounds, as it provides a direct functional relationship between parameters deemed most important by the user. 
\end{itemize}
Code for environments is open source at the Interactive Robotics Laboratory gym \cite{beard2022irl_gym} and algorithms can be found in our decision making toolbox \cite{beard2022dm}.

\subsection{Paper Structure}

The remainder of the paper is as follows. 
Section \ref{sec:amdp} contains details regarding decision making under ambiguity: defining ambiguous MDP's and belief functions.  
Section \ref{sec:methods} describes the approach used to generate belief functions, as well as the AAGS algorithm.
Section \ref{sec:results} discusses results.
Lastly, in Section \ref{sec:conclusion}, the reader will find conclusions and future work. 

\section{Ambiguity MDP} \label{sec:amdp}

Markov decision processes (MDP) have seen broad use formulating decision problems \cite{kochenderfer2015decision} and have a substantial base of literature from which to build tools. 
The key issue with MDPs is the assumption of one true transition model. For methods which produce approximate solutions, this model is hidden.
Depending on the approach taken, then, multiple transition models may be inferred, implying significant ambiguity when data is limited. 
This work builds on the concept ambiguous Markov decision processes (AMDP) \cite{saghafian2018ambiguous, bauerle2019markov}, which frame the problem with a set of transition models.
Define the AMDP as $\mathcal{M} \coloneqq \langle \mathcal{S}, \mathcal{A}, \mathcal{T}, R, \gamma \rangle$. 
Let $s \in \mathcal{S}$ be a state in the set of states and $a \in \mathcal{A}$ be an action in the action space. 
The reward $R(s,a,s')$ defines the optimization criterion with respect to a specified state transition. 
The variable $\gamma$ acts a temporal discount factor. 
Lastly and central to the AMDP formulation, $T(s'|s,a) \in \mathcal{T}$ is a model among the set valid of transition models. In this work the set of transition models is represented as a special class of random sets\textemdash referred to as belief functions (different from beliefs in POMDP literature\footnote{Beliefs are defined specifically with respect to state uncertainty. Belief functions refer to mass functions whose elements may support more than one outcome.}).  
As opposed to previous AMDP formulations, use of belief functions are better suited to discrete distributions than cloud models as used by Saghafian \cite{saghafian2018ambiguous}. Additionally, these belief functions generalize Bayesian mass functions, allocating mass to multiple hypotheses \cite{cuzzolin2020geometry}. Thus they permit more comprehensive expressions of ambiguity than those derived from Bayesian models (\textit{e.g.}, the AMDP of Ba{\"u}rle and Rieder \cite{bauerle2019markov}).

\begin{figure}[!t]
    \centering
    \smallskip
    \begin{subfigure}{0.47\linewidth}
        \centering
        \includegraphics[width=\linewidth]{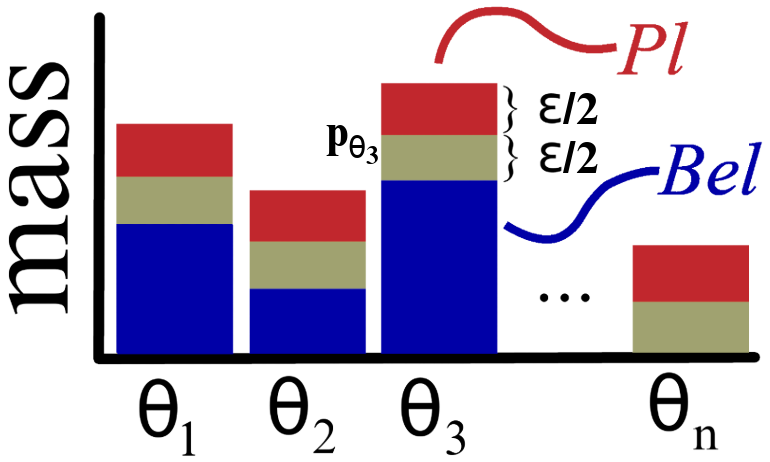}
        \caption{}
    \end{subfigure}%
    \begin{subfigure}{0.47\linewidth}
        \centering
        \includegraphics[width=\linewidth]{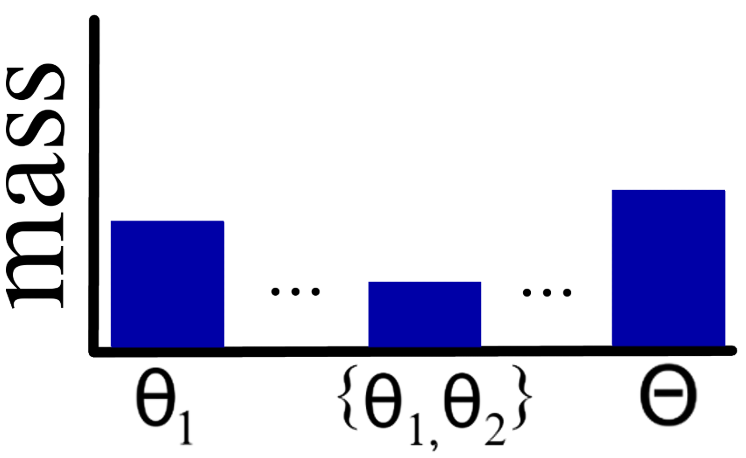}
        \caption{}
    \end{subfigure}
    \caption{Our approach accepts multinomial distributions estimated by collecting samples from a blackbox simulator. (a) Using confidence intervals of size $\epsilon$ for each mass term, we constrain the model. This yields upper $Pl$ and lower $Bel$ estimates on the mass of each outcome. (b) Using these bounds, we distribute the mass to a belief function. Such a formulation assigns mass to multiple valid distributions when information is ambiguous.}
    \label{fig:mass_dist}
\end{figure}

To define a set of transition models with a belief function, let $\mathcal{S}' \subseteq \mathcal{S}$ be the space of outcomes from some state-action pair $(s,a)$. 
Then distribute mass to the power set of outcomes $2^{\mathcal{S}'}$, such that the mass of some proposition $m(s'): 2^{\mathcal{S}'} \rightarrow [0,1]$ and $\sum_{s'} m(s') = 1$, $ \forall s' \in \mathcal{S}'$ \cite{cuzzolin2020geometry}. 
As an example let our belief function over $\mathcal{S}' = \{s_1,s_2\}$ be such that $m(\{s_1,s_2\})=0.6$, $m(\{s_1\})=0.1$, and $m(\{s_2\})=0.3$. 
For the latter two, mass acts as is typical of a standard probability.
For $m(\{s_1,s_2\})$, this posits that, given our limited knowledge, 60\% of the time we would be unable to predict the outcome. 
These belief functions are typically mapped to lower and upper probabilities rather than using the undistributed mass directly. 
Let there be two sets $\mathcal{B}, \mathcal{C} \subset \mathcal{S}'$
The lower probability or belief represents the minimum mass supporting a set of outcomes
\begin{equation}
    Bel(\mathcal{B}) = \sum_{\mathcal{C} \subseteq \mathcal{B}} m(\mathcal{C}),
\end{equation}
while the upper probability or plausibility represents corresponding maximum
\begin{equation}
    Pl(\mathcal{B}) = \sum_{\mathcal{B} \cap \mathcal{C} \neq \emptyset} m(\mathcal{C}).
\end{equation}
These bounds on probability are demonstrated in Figure \ref{fig:mass_dist}.a where each mass is known to $\pm \epsilon/2$. Figure \ref{fig:mass_dist}.b captures the distribution as a belief function where mass can now support multiple hypotheses.

With a set of transition models, it is no longer possible to use Bellman's equation and the expected value to solve the AMDP. 
The $\alpha$-Hurwicz function is a common choice as it allows for models with varying ambiguity attitudes \cite{arrow1972optimality}.
While there are several alternatives \cite{ghirardato2004differentiating} (including work which integrates expected value, risk, and ambiguity into a single metric \cite{he2019additive}),  $\alpha$-Hurwicz has some the desirable properties explored at greater length in Sec. \ref{sec:methods_belief}.
According to $\alpha$-Hurwicz, the value is defined as follows 
\begin{equation} \label{eq:alpha_criterion}
    V(s) = \max_{a}[ (1-\alpha)\underline{\mathbb{E}}(s,a) +  \alpha\overline{\mathbb{E}}(s,a)].
\end{equation}
Here the ambiguity attitude $\alpha \in [0,1]$ controls how optimistically the value is perceived.
The lower expectation is 
\begin{equation}
    \underline{\mathbb{E}}(s,a) = min_{T \in \mathcal{T}} \sum_{s'}T(s'|s,a)[(r(s,a,s') + \gamma \underline{\mathbb{E}}(s',a^*)].
\end{equation}
The upper expectation similarly follows suit, with the $max$ function.
Note that $a^*$ indicates the optimal action for $s'$.

\section{Methods} \label{sec:methods}

The goal of this work is to motivate the broader study of ambiguity in decision making. 
The proposed ambiguity attitude graph search algorithm builds on this by better representing ambiguity present in decision making, and generalizing robust decision making to include ambiguity attitudes. 
As with many sampling based methods \cite{kocsis2006bandit, leurent2020monte, kochenderfer2015decision}, agents supply an $(s,a)$-pair to a simulator. This simulator returns observations in the form of a state transition and reward $\omega \coloneqq (s',r)$.
Using these observations, the agent constructs an estimate of the transition model for each $(s,a)$-pair.
However, multiple underlying transition models could have produced these observations. Lacking additional information, there is ambiguity and the model estimate should instead be a set of models. Whereas prior methods, such as graph based optimistic planning (GBOP) \cite{leurent2020monte}, assume ambiguity is adequately described by confidence intervals, we contribute a novel way to generate belief functions from these intervals to more comprehensively represent the distribution of mass. While robust methods assume the lower bound should be used to arbitrate models, AAGS uses ambiguity attitudes to balance robustness and optimism. 



\subsection{Computing Belief Functions}\label{sec:methods_belief}

Approximate decision making algorithms generally define their optimality either with respect to the budget or accuracy of estimated models \cite{jonsson2020planning}. 
For fixed confidence approaches, a confidence interval over some value is specified to accuracy $\epsilon/2$ given confidence $\delta$.
The idea is that a quantity $z$ is approximated with $\hat{z}$ such that $P(|z - \hat{z}| < \epsilon/2) \geq 1 - \delta$.
Whereas these bounds are generally applied to the reward or value estimates, AAGS bounds the mass terms to evaluate the distribution more directly.
Focusing on the entire distribution in this way has been shown to be more expressive and provide a better understanding of the model \cite{bellemare2017distributional}.

When there are limited samples, it is unlikely the constraints for a fixed confidence approach will be satisfied. 
We therefore develop a ``floating confidence" description of the set of transition models for this case. Thus users can understand the confidence with which a distribution of \textit{desired} accuracy may be interpreted. 
A belief function is then evaluated using linear optimization.
Using the confidence, the inferred set of models is discounted against the set of all models to arrive at a distribution expressing the ambiguity. 

Begin by assuming the transition model is represented by a multinomial distribution, known to $(\epsilon,\delta)$.
Our approach is agnostic to how these are obtained (for clarity, implementation details are presented in the Appendix).
We wish to redistribute the $\epsilon$ mass from each outcome to ambiguous mass terms, generating a set of transition models. 
Define $\Theta$ as the set of outcomes $(s',r)$ from our state transition, and let $2^\Theta$ be the power set of all such outcomes.
Let each proposition of size one be $\omega_i \in \Theta$ and propositions of size two or more be $\theta_j \in \{2^\Theta \setminus \Theta\}$. Let $p_w$ be the sampled probability of some $\omega$. 
The accuracy constrains belief functions such that the lower and upper bounds are, respectively, $Bel(\omega)=p_\omega-\epsilon/2$ and $Pl(\omega) = p_\omega + \epsilon/2$, $\forall \omega \in \Theta$.
Because these are probabilities, they are clipped to 0 and 1, as necessary.

Using these constraints, the problem is structured as a linear optimization $Ax=B$.
Note, for each proposition $\omega$,
\begin{equation} \label{eq:mass_dist}
    Pl(\omega)-Bel(\omega) =\sum_{\theta \ni \omega} m(\theta).
\end{equation}
Let the columns $j$ of $A$ correspond to each $\theta_j$ and each row $i$ to the singleton outcomes $\omega_i$.
Then compose the $A$ matrix using the indicator function $\mathds{1}(\omega_i) \coloneqq \omega_i \in \theta_j$.
From Eq. \ref{eq:mass_dist}, each row of $B$ is equal to $Pl(\omega_i)-Bel(\omega_i)$.
Next append a row of ones to the end of $A$ and $1-\sum Bel(\omega)$ to $B$ to constrain the total mass by that available.
Then solve for the column $x$ of all $\theta_j$, as these are the unknown ambiguous mass terms.
This optimization problem is as follows:
\begin{equation} 
    \begin{bmatrix}
        \multicolumn{4}{c}{\mathds{1}(\omega_i)} \\
        1 & 1 & ... & 1
    \end{bmatrix}
    \begin{bmatrix}
        2^\Theta \setminus \Theta \\
    \end{bmatrix}
    =
    \begin{bmatrix}
        Pl(\omega_i)-Bel(\omega_i) \\
        1-\sum Bel(\omega_i)
    \end{bmatrix}.
    \label{eq:lin_eq}
\end{equation}
In practice the number of terms in $2^\Theta$ can become intractable. 
We arbitrarily limited $|\Theta|\leq 12$ as this exceeded the branching factor for our experiments. Then $A^{-1}$ was precomputed to save resources online.

Because accuracy is fixed when there are few samples, confidence will be low for transition models. Therefore it is likely the true model lies somewhere in the entire set of transitions. This information is factored into the estimate using the discount rule of Shafer \cite{shafer1976mathematical} with the confidence as the discount parameter. However, the set of models is unknown. Choice of the $\alpha$-Hurwicz decision criterion overcomes this issue by loosening the requirement to knowledge of the reward bounds. 
\begin{lemma}
    Let $U \cup L \cup \Theta$ be the entire outcome space of the decision, where $U$ has a reward equal to the upper bound of the outcome space and $L$ the lower bound.
    For a given belief function $b$ over $U \cup L \cup \Theta$, inclusion of any other element $\omega= (s',r) : L \leq r \leq U$ does not change the upper $\overline{\mathbb{E}}(b)$ or lower $\underline{\mathbb{E}}(b)$ expectations.
\end{lemma}

\noindent\textit{Proof.} We begin by looking at the upper expectation; the lower expectation follows trivially. By definition of the upper expectation, the mass goes to the element in each set having the greatest magnitude. All $r$ less than $U$ result in all mass allocated to $U$. If $r = U$, the terms are the same and we are left with the original set $\{U \cup L \cup \Theta\}$ of outcomes. \null\nobreak\hfill\ensuremath{\square}

Let $m_b$ be the mass of elements in our belief function.
Thus, the original model to $(\epsilon,\delta)$ accuracy/confidence is equivalent to the following discounted mass function $\{m(\theta) = (1-\delta)m_b(\theta)$ for $\theta \in 2^\Theta$, $m(\{U \cup L \cup \Theta\}) = \delta\}$.
For simplicity, let $\theta$ now include elements of size one.
This is reflected in the Algorithm \ref{alg:dist2belief}; here $P$ is our sampled distribution.
Notice in line two, the model confidence is computed from the accuracy and number of samples used to make the distribution $P.N$.
This is used to generate the constraints $Bel,Pl$ on the model.

\begin{algorithm}

$b \leftarrow \{\}$, $n = size(P)$

$Bel, Pl \leftarrow f(\epsilon,P.N)$

$b \leftarrow b \cup (1-\delta)Bel$

$m = A^{-1}B \qquad$ \textit{from Eq. \ref{eq:lin_eq}}

\For{$\theta \in 2^{P.\Theta}$}
{
    $b \leftarrow b \cup (\theta,(1-\delta) m(\theta))$
}

return $b \cup (\{U \cup L \cup P.\Theta\}, \delta)$

\caption{dist2belief($P$,$\delta$,$U$,$L$)} \label{alg:dist2belief}
    \SetAlgoLined
\end{algorithm}

\subsection{Ambiguity Attitude Graph Search}

The algorithm ambiguity attitude graph search (Alg. \ref{alg:AAGS}), generalizes robust planners to permit ambiguity attitudes. 
These attitudes are specified by $\alpha \in [0,1]$, where $\alpha=0$ is a conservative or a robust policy.
Conversely, $\alpha = 1$ is an optimistic or ambiguity seeking policy.
We assume the user knows the reward bounds $R_{min}$, $R_{max}$.
For an infinite horizon MDP, the bounds on the value follow as $R_{max}/(1-\gamma)$ and $R_{min}/(1-\gamma)$ \cite{leurent2020monte}.
Users also specify accuracy and confidence requirements $\epsilon$, $\delta$.

The AAGS algorithm is a graph based planner; when a state is reached multiple times, rather than creating a new node, the existing node is added as a child of the current node. In doing so, the graph using samples more efficiently than tree methods\textemdash as the decision maker can leverage more prior knowledge \cite{leurent2020monte}.
Assuming the simulator is much more expensive than the algorithm itself, the cost of any added computation are far outweighed by the reduced sample complexity.
Cyclic graphs can occur during search and there are no constraints on connectivity. 

\begin{algorithm}
\SetKwProg{Proc}{Procedure}{}{}
\SetKw{Input}{input:}
\SetKw{Initialize}{initialize:}

\Input{$\alpha, R_{max}, R_{min}, \epsilon, \delta, N, D$}

\Initialize{$\mathcal{G} \leftarrow \{\}$}

\Proc{search($s_0$)}
{
\If{$s_0 \not\in \mathcal{G}$}
{
    $\mathcal{G} \leftarrow \mathcal{G} \cup s_0$, 
}
$n \leftarrow 0$

\While{$n < N$}
{
    $s \leftarrow s_0$
    
    $parents \leftarrow s$
    
    \While{$s \neq terminal$ and $depth < D$}
    {
        
        $a \leftarrow \arg\max_b \overline{\mathbb{E}}(s_0,b)$
        
        \If{$t(s,a) < t(\epsilon,\delta) $}
        {
            $s', r \sim G(s,a)$
        }
        \Else
        {
            $s', r \sim T(s,a)$
        }
        
        $\mathcal{G}(s,a).update(s', r)$
        
        $parents \leftarrow parents  \cup s'$, $s \leftarrow s'$
    }
    
    $backpropagate(parents)$
}
    
$return \arg\max_a \alpha \underline{\mathbb{E}}(s_0,a) + (1-\alpha) \overline{\mathbb{E}}(s_0,a)$
}

\Proc{backpropagate($P$)}
{
    \For{$s \in parents$}
    {
    $parents.pop(s)$
    
    $L \leftarrow s.L$, $U \leftarrow s.U$

    $a, s.V \leftarrow \max_a \alpha \underline{\mathbb{E}}(s,a) + (1-\alpha) \overline{\mathbb{E}}(s,a)$
    
    $s.L \leftarrow \underline{\mathbb{E}}(s,a)$, $s.U \leftarrow \overline{\mathbb{E}}(s,a)$
    
    \If{$s.L - L > \beta$ or $U - s.U > \beta$}
    {
        $parents \leftarrow parents \cup s.parents$
    }
    }
}
\caption{Ambiguity Attitude Graph Search} \label{alg:AAGS}
    \SetAlgoLined
\end{algorithm}

AAGS begins from state $s_0$ with an empty graph $\mathcal{G}$ and searches the solution space with $N$ trajectories up to horizon $D$. 
For each time step, the action having the greatest upper expectation is selected (in line with the optimism in the face of uncertainty paradigm \cite{leurent2020monte}).
Both the upper and lower expectations compute the belief functions from Alg. \ref{alg:dist2belief} internally. 
If a state has not been sampled enough to be considered known (\textit{i.e.}, satisfying the constraints $\epsilon$, $\delta$ for all actions), the generative model $G$ is sampled.
Otherwise, we borrow from the ``knows what it knows" framework \cite{li2011knows} and sample the estimated model $T$ to reduce computational overhead.
New outcomes are added to the graph and the current state is added to the list of parents.
Next, our bounds are backpropagated following the approach of Leurent, \textit{et al.} \cite{leurent2020monte}.
Therein, upper and lower bounds are updated through an approximate value iteration over the set of parents.
Parents are appended to the list for those states, whose bounds have changed by more than $\beta = \dfrac{1-\gamma}{\gamma}\Delta$.
Here $\Delta$ is the regret or difference in value between the second best and optimal actions for our model at a state. 
This differs with respect to the work of Leurent, \textit{et al.} in that we compute upper $s.U$ and lower $s.L$ value bounds at a state through belief functions rather than confidence intervals. 
More importantly, we differ by using Eq. \ref{eq:alpha_criterion} to select the best action\textemdash both in backpropagate and when selecting the policy. This incorporates the ambiguity attitude $\alpha$ both into the final decision and propagates the value estimate through the entire tree to ensure consistency.

\section{Results} \label{sec:results}

We compare AAGS against two algorithms. First, UCT \cite{kocsis2006bandit} serves as a baseline. Because UCT does not consider sets of models in its value estimates, it is ambiguity neutral. 
UCT was tuned to have a rollout horizon of 25 time steps and exploration parameter $c=8$.
Next, GBOP \cite{leurent2020monte} was used to compare against robust planning. Both AAGS and GBOP were given a horizon of 50 time steps. All algorithms were allocated a budget of 500 samples from the generative model.

\subsection{Expression of Ambiguity} \label{sec:eoa}

First, let us draw attention to the parameter $c$ used by UCT. 
This exploration parameter affects its search, not the representation of value during decision making.
Thus, at leaf nodes there is no concept of uncertainty in the backpropagated reward. 
Robust algorithms have begun to consider this uncertainty, however, they assume the lower bound is most relevant to the decision. 
This motivates the use of ambiguity attitudes to reflect how willing the agent should be to accept unknown gambles.
One impact, is that users can control the extent to which the agent explores its environment. 
This is demonstrated in Figure \ref{fig:tunnel_top} where we compare the role of planning horizons with ambiguity attitude for AAGS. 
In some cases long planning horizons ($h = 100$) are necessary to for a robust attitude ($\alpha = 0$) to identify high value targets. In which case, they move directly towards the target.
Given a short planning horizon and an increasingly optimistic attitude ($h = 10;\; \alpha \rightarrow 1$), however, agents increasingly seek alternatives. They move towards ambiguous regions to learn about their environment. This results in more exploratory behavior despite shorter search horizons. 

\begin{figure}[t!]
    \centering
    \centering
    \vspace{5px}
    \includegraphics[width=0.85\linewidth,trim={3px 0 0 143px},clip]{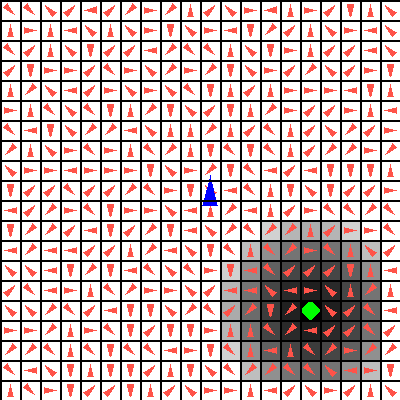}
    \caption{Example of a sailing world environment; the agent (blue triangle) aims to reach the goal in green and is increasingly rewarded (darker regions) as it gets closer to the goal (green). Concurrently, the agent must try to avoid boundaries and follow the wind (red triangles).}
    \label{fig:envs}
\end{figure}

\subsection{Sailing}

To further illustrate the role of ambiguity in decision making, consider the sailing environment Leurent, \textit{et al.} \cite{leurent2020monte} used to evaluate GBOP (Fig. \ref{fig:envs}). We modify the environment so the agent has a broken rudder and can only turn 0, 45, or 90$^\circ$ to increase the difficulty. While attempting to reach the goal, the agent must avoid reefs at the boundary which incur significant damage. At each cell is a wind vector, with probability $w=10\%$ of changing direction by $\pm45^\circ$. Because the wind vector in all cells may change, the chances a given state will be encountered more than once is near zero.  

From the perspective of the decision making tool, then, there will be significant ambiguity. For an $m$-by-$n$ grid, nodes in the graph will be faced with $3^{mn+1}$ transitions, but never sample more than one. Thus we can simulate ambiguity in a relatively compact environment. Because the state is large and the problem is fully observable, search algorithms become quite expensive. We therefore limit the environment to a 40-by-40 grid with 500 trials of random state-goal pairs. 

\begin{figure}[!t]
    \centering
    \vspace{5px}
    \begin{subfigure}{0.47\linewidth}
        \centering
        \includegraphics[width=\linewidth]{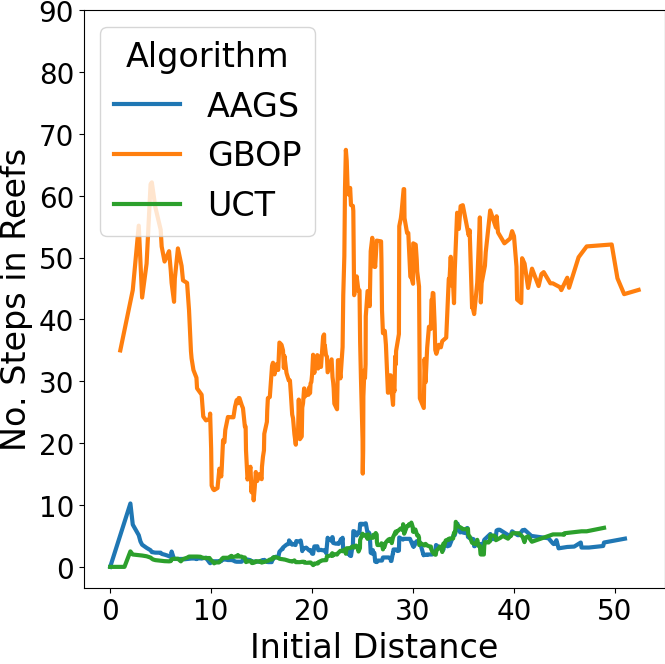}
        \caption{}
    \end{subfigure}%
    \begin{subfigure}{0.47\linewidth}
        \centering
        \includegraphics[width=\linewidth]{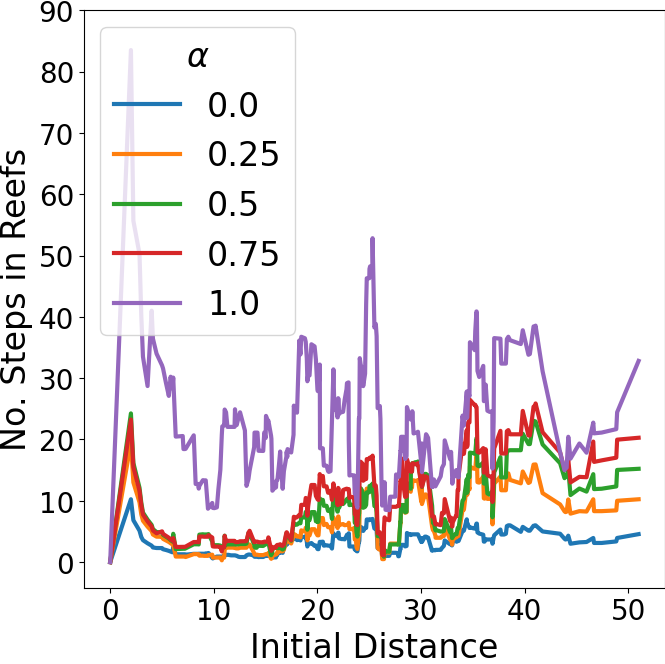}
        \caption{}
    \end{subfigure}

    \caption{(a)The average steps in the reefs for UCT and robust solutions (AAGS at $\alpha = 0$ and GBOP) and (b) the average steps AAGS spent in reefs for different $\alpha$ values. GBOP often became mired in the reefs. However, UCT and robust AAGS actively avoided them. By increasing $\alpha$, AAGS is incrementally more likely to spend time in reefs. }
    \label{fig:robustness}
\end{figure}

\begin{figure}[!t]
    \centering
    \medskip
    \begin{subfigure}{0.47\linewidth}
        \centering
        \includegraphics[width=\linewidth]{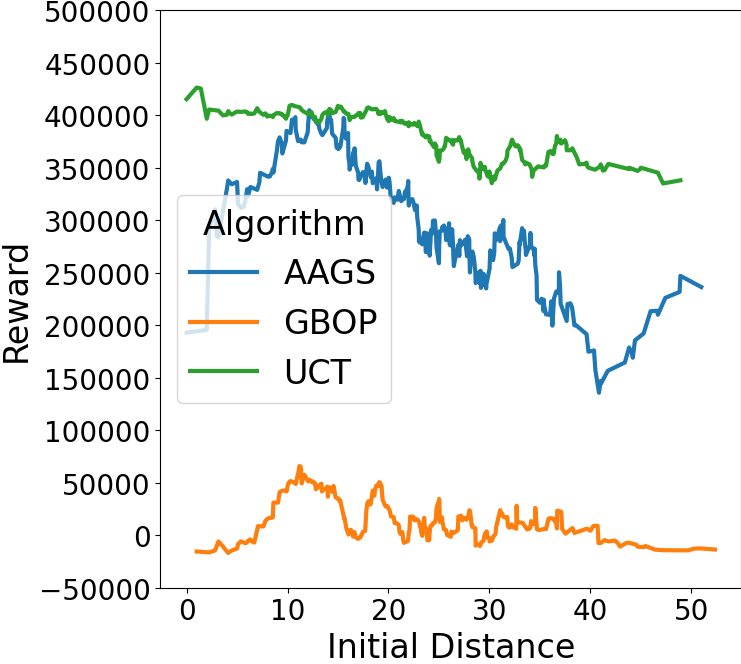}
        \caption{}
    \end{subfigure}%
    \begin{subfigure}{0.47\linewidth}
        \centering
        \includegraphics[width=\linewidth]{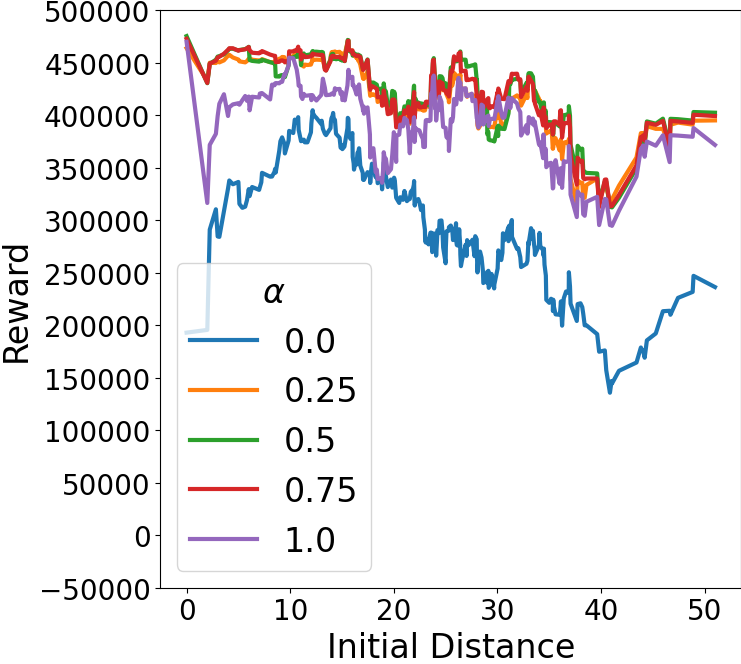}
        \caption{}
    \end{subfigure}
    \caption{(a) Reward for UCT and robust solutions (AAGS at $\alpha = 0$ and GBOP) and (b) reward achieved by AAGS by varying $\alpha$. Focusing on raising the lower bound, robust attitudes achieve lower reward. Conversely, slightly positive attitudes can outperform ambiguity neutral methods (UCT). When attitudes are too optimistic ($\alpha=1$), unnecessary gambles harm performance. }
    \label{fig:s_r_d_all}
\end{figure}

We first look to the role of ambiguity in avoiding unsafe outcomes. Interestingly GBOP, a robust algorithm struggles with this (Fig. \ref{fig:robustness}.a). We postulate this is because the fixed confidence models employed are relatively uninformative in environments with high entropy transitions such as this. The UCT and robust ($\alpha=0$) AAGS algorithms, however, almost completely avoid such negative outcomes. Though this number does trend up slightly with increases in initial distance (from the goal), as agents are more likely to start in the boundary. Furthermore, we see that by adjusting the ambiguity attitude of AAGS, we can directly control the willingness of an agent to gamble on low quality outcomes (Fig. \ref{fig:robustness}.b). This is further reflected by the rewards (Fig. \ref{fig:s_r_d_all}). For moderate attitudes, an agent can effectively balance reward seeking with chances of hitting the reefs. This can lead to improvements over UCT by $\sim$10\%. Conversely, a robust attitude may significantly underperform. Similarly, an optimistic one may gamble too much and degrade outcomes.

\section{Conclusions} \label{sec:conclusion}

The increasing complexity of tasks we assign to robots entails ambiguity on the part of the decision maker. 
This work explores how a decision maker \textit{infers} multiple models. While existing methods do this in varying capacities, this work aims to demonstrate how more explicit representations of ambiguity contribute to understanding uncertain decisions. 
In doing so, this lays the groundwork for novel multiple model problems. 
Additionally, this work demonstrates how ambiguity attitudes, considering both robust and optimistic outcomes, can lead to better control over the balance between safety and improving outcomes. 
By generalizing robust decisions, tools like AAGS can be used beyond safety critical problems. Future work should focus on formal guarantees for problems with optimistic ambiguity attitudes. Furthermore, building on the work of Marinacci \cite{marinacci2015model}, could lead to methods which independently control how agents consider both risk and ambiguity.


\section*{ACKNOWLEDGMENT}

This research was made possible by the NASA West Virginia Space Grant Consortium, Grant \#80NSSC20M0055.

\bibliographystyle{IEEEtran} 
\bibliography{root}

\section*{APPENDIX}

Confidence intervals are used in this work: to compute confidence given a desired accuracy and the number of samples, as well as, to use the accuracy and confidence requirements to specify a rule to stop sampling the simulator. 
As there were no existing relations, Monte Carlo trials were used to approximate one by sampling multinomial distributions:
\begin{equation}
    t(\epsilon,\delta) = \dfrac{ln\dfrac{1}{1.25(1-\delta)-1/6}}{\epsilon(-ln\dfrac{1}{1.5(1-\epsilon)+1/3})^2}.
    \label{eq:samples}
\end{equation}
As intuition would suggest, the following constraints are also satisfied.
First, $\delta(\epsilon=1,t) = 0$: no requirement on accuracy implies it is met trivially.
Next, $\epsilon(\delta=0,t) = 0$: no confidence in a model implies it is known to perfect precision.
This approximation performed well and better matched the distribution than existing (algorithm agnostic) bounds \cite{li2011knows,kakade2003sample}, while constraining $\epsilon,\delta \in [0,1]$.
To solve for $\epsilon$, we used gradient descent.



\addtolength{\textheight}{-12cm}   


\end{document}